# MIST: Medical Image Segmentation Transformer with Convolutional Attention Mixing (CAM) Decoder

Md Motiur Rahman  Shiva Shokouhmand  Smriti Bhatt  Miad Faezipour

Purdue University

West Lafayette, IN 47907, USA

`{rahma112, sshokouh, bhatt32, mfaezipo}`@purdue.edu

## Abstract

*One of the common and promising deep learning approaches used for medical image segmentation is transformers, as they can capture long-range dependencies among the pixels by utilizing self-attention. Despite being successful in medical image segmentation, transformers face limitations in capturing local contexts of pixels in multimodal dimensions. We propose a Medical Image Segmentation Transformer (MIST) incorporating a novel Convolutional Attention Mixing (CAM) decoder to address this issue. MIST has two parts- a pre-trained multi-axis vision transformer (MaxViT) is used as an encoder, and the encoded feature representation is passed through the CAM decoder for segmenting the images. In the CAM decoder, an attention-mixer combining multi-head self-attention, spatial attention, and squeeze and excitation attention modules is introduced to capture long-range dependencies in all spatial dimensions. Moreover, to enhance spatial information gain, deep and shallow convolutions are used for feature extraction and receptive field expansion, respectively. The integration of low-level and high-level features from different network stages is enabled by skip connection, allowing MIST to suppress unnecessary information. The experiments show that our MIST transformer with CAM decoder outperforms the state-of-the-art models specifically designed for medical image segmentation on the ACDC and Synapse datasets. Our results also demonstrate that adding the CAM decoder with a hierarchical transformer improves the segmentation performance significantly. Our model with data and code is publicly available on GitHub[1].*

## 1. Introduction

Medical image segmentation is necessary to diagnose diseases, formulate treatment plans, and assess treatments. It is one of the essential steps for providing treatments through computer-aided systems [22, 25]. By performing precise segmentation of medical images, computer-aided systems assist physicians and facilitate the identification of potential tumors, localization of lesion boundaries, and contribute to accurate diagnoses [27]. Precise segmentation also expedites diagnosis and improves tumor detection rates. Initially, the convolutional neural network (CNN)-based symmetric encoder-decoder architecture such as UNet [28] was found promising for medical image segmentation, as its encoder part compresses the image into a latent features space which carries more encoded information of the image. The decoder part learns and finds the regions of interest (ROI) by expanding the latent features and utilizing the multi-stage feature maps from the encoder through skip connections [17]. Along with the U-shaped UNet encoder-decoder architecture with horizontal skip connections that has become so popular for medical image segmentation due to capturing local context more precisely and generating high-resolution segmentation maps, some other variants of UNet like UNet++ [44], DC-UNet [24], and UNet 3+ [11] were developed for this purpose. While CNN-based encoder-decoder architectures have shown success in medical image segmentation, the convolution operator in CNN-based encoder-decoder architectures can only capture local context. Limitations exist in extracting long-range dependencies among image pixels. Numerous studies have tackled this matter by integrating attention mechanisms in CNN-based models [41, 40]. It has been observed that attention-based CNN models outperform the sole CNN-based models in the context of medical image segmentation. Additionally, other researchers have utilized dilated convolution to expand the receptive field of kernels [8, 19], enabling the CNN models to capture long-range dependencies better. Despite the improvements seen in these CNN models with added mechanisms, more than merely the resulting global contexts is needed to showcase long-range dependencies effectively.

To address these limitations, the vision transformer (ViT) [7], a variation of the original transformer mainly developed for natural language processing (NLP) tasks, was found more successful, as it can capture the long-range dependencies more effectively through self-attention mechanisms. The ViT generates patches, known as tokens in NLP tasks, and applies self-attention to capture correlation among all the patches, enabling the ViT to capture more long-term dependencies. Due to the significant success of ViT in medical image segmentation, several variations of the transformer, such as Swin transformer [23], Pyramid vision transformer (PVT) [33], and Convolutional vision transformer (CvT) [35], have been developed for medical image segmentation. Although they overpower the CNN-based models in terms of performance in medical image segmentation, they have limited capabilities of capturing local contexts. This issue is addressed in SegFormer [36], and UFormer [34] by adding convolution layers in the transformer, allowing them to capture the local and global contexts together.

---

[1] https://github.com/Rahman-Motiur/MIST

Despite the ability to capture both local and global contexts in recent techniques, the attention mechanism needs to be improved to capture enough short and long-range dependencies of each pixel in all spatial dimensions. Considering this limitation, we develop a new Medical Image Segmentation Transformer (MIST) by combining a Multi-Axis Vision Transformer (MaxViT) as an encoder and a proposed novel decoder named Convolutional Attention Mixing (CAM). We propose a unique decoder architecture combining self-attention, spatial attention, and squeeze and excitation attention. The main contributions of this study are as follows:

- We propose a new Convolutional Attention Mixing (CAM) Decoder designed explicitly for segmenting medical images. Each block of the CAM decoder is built based on an attention-mixer that combines computed multi-head self-attention, spatial attention, and squeeze and excitation attention modules. This allows the decoder to take advantage of all the dependencies among the image pixels during segmentation. We also introduce convolutional projection instead of linear projection during self-attention estimation, which reduces computational costs and addresses more spatial information. By doing so, the model can learn the essential regions of the image with better precision and determine the attention weight of each pixel in all spatial dimensions. This technology allows the model to converge faster and provide accurate segmentation results.
- To enhance the model's ability in learning significant latent features, we employ skip connections that leverage the encoder's multi-stage features and merge them with the up-sampled decoder features at multiple levels. In our approach, we incorporate deep and shallow convolutions with varying dilations to expand the model's kernel receptive fields. This enables capturing more long-term dependencies among image pixels and extracting pertinent features. The model gains more useful spatial information by combining the features from multi-dilated convolutions.
- We conduct various ablation studies to demonstrate the effectiveness of our CAM decoder in segmenting 2D medical images and evaluate the impact of different decoder components.

The rest of the paper is organized as follows: related works are discussed in Section 2; the model and its implementation are illustrated in Section 3; Section 4 illustrates the obtained results; and Section 5 concludes the paper.

## 2. Related Work

We divide and discuss the related works in three subsections: CNN-based models and attention, vision transformers, and CNN-transformers for medical image segmentation.

### 2.1. CNN-based Models and Attention

The UNet [28] model is the first CNN-based model applied to medical image segmentation. Its impressive performance led to developing other UNet variants, including UNet++ [43]. UNet++ utilizes nested and dense skip connections to connect the encoder and decoder, effectively reducing the gaps between feature maps. Subsequent variants of UNet++ [44] have been proposed by redesigning the skip connections to enable flexible fusion of feature maps, resulting in improved performance compared to the earlier UNet++ model. The limitations of UNet++ were addressed in another study that proposed UNet3+ [11] which leverages deep supervision and full-scale skip connections. However, these models have limited capabilities in capturing long-range independencies, leading to the popularity of attention-based UNet models in medical image segmentation. Some examples include the MA-Unet [2], which uses the attention mechanism with multi-scale feature utilization, and APAUNet [15], which uses axis projection attention, performing better than UNet models without attention mechanisms. Other attention mechanisms, such as spatial and residual attention, are also used in many UNet models. Spatial channel attention is introduced in the SCAU-Net [41] model, while residual attention is used in Residual-Attention UNet++ [21]. One of the limitations of UNet models is adaptation which is addressed in the nnU-Net model [14], with a self-adaptation capability to pre-process data and generate the optimal model for better prediction.

### 2.2. Transformers for Medical Image Segmentation

Transformers have significantly impacted natural language processing and are now being developed for computer vision tasks, particularly for segmenting medical images. Vision transformers (ViT) [7] are found to be more efficient than CNN models for medical image segmentation. The development of various ViT variants has resulted in further improved performance while reducing computational costs. For example, the Swim transformer [23] uses sliding-window-based attention to reduce costs and improve performance. The SegFormer [36] transformer replaces the initial ViT's positional encoding and simplifies the decoder with a multi-layer perceptron (MLP)-based approach. However, transformers cannot capture local spatial features. This issue is addressed in the Convolutional vision transformer (CvT) [35] by using convolution in the original ViT. The nnFormer [42] is another convolutional transformer that utilizes global and local volume-based self-attentions to learn better volumetric representations.

The UNet shape encoder-decoder is commonly used in vision transformers. Popular transformers for medical image segmentation include the Swin Unet3D [1], Uformer [34], U-Net Transformer [25], and TransClaw U-Net [4], which all follow a U-shaped encoder-decoder architecture. Attention mechanisms are critical to transformer operation, and different forms of attention have been introduced to extract local and global contexts. For example, the TransAttUnet transformer [5] uses self-aware and global spatial attention to extract non-local interactions, while the U-shaped transformer [39] uses neighborhood attention to capture local context. The MaxViT transformer [29] introduces squeeze and excitation attention, and the medical transformer [30] uses gated axial attention. Combining transformers with UNet decoders has also proven successful in medical image segmentation. The LeViT UNet [37] and MaxViT UNet [16] models combine LeViT and MaxViT transformers with a UNet model, and have outperformed sole transformers regarding accuracy.

Various hierarchical transformers utilizing multi-scaled features were used as backbones for medical image segmentation. HiFormer [10] is a recently proposed hierarchical transformer that bridges a CNN model with the seminal Swin transformer and fuses the representations from these two using double-level fusion. Another commonly used backbone is pyramid vision transformer (PVT) [33] which reduces the computational costs of processing high-resolution feature maps by shrinking them to low-resolution feature maps.

### 2.3. CNN-Transformers for Medical Image Segmentation

In the segmentation field, hybrid transformers that combine CNN and transformers have been gaining popularity recently. One example is the ConvTransSeg model [9], which utilizes CNN as the encoder and transformer as the decoder. Another model called Cats [18] also combines a CNN and transformer as encoders. Fusing these two models can extract local and global contexts more effectively. TFCNs [20] is another hybrid transformer that utilizes dense convolution and convolution linear attention to capture better semantic segmentation features.

Although all these different variants of vision transformers perform well in medical image segmentation, they face some limitations, such as the linear nature of the attention mechanism in transformers which increases the computational costs, and not being able to capture the spatial contexts. In addition, the existing models for medical image segmentation have limited capabilities capturing short and long-range dependencies in all multi-modal spatial dimensions. In this study, the proposed MIST transformer with the CAM decoder alleviates all these issues. We use an attention-mixing strategy where we combine convolutional projected self-attention, spatial attention, and squeeze and excitation attention computed at different stages of the decoder block.

## 3. Method

Our Medical Image Segmentation Transformer (MIST) consists of a Multi-Axis Vision Transformer (MaxViT) used as an encoder combined with a proposed novel decoder named Convolutional Attention Mixing (CAM) decoder. The complete architecture of our model is discussed in this section.

### 3.1. Convolutional Attention Mixing (CAM) Decoder

The CAM decoder has four consecutive decoder blocks and one bottleneck block. The right side of Figure 1 shows the structure of our proposed CAM decoder. Each decoder block starts with *layer-normalization*, and then up-sampling of the inputs by a factor of two to match the size of the encoder outputs. Next, we apply *convolution* followed by *layer-normalization* by keeping the input dimension the same and concatenating with the encoder output through the skip connection, as presented in Equation 1.

$$X = Cat(N(C(U(N(X)))), Enc) \quad (1)$$

where $Enc$ refers to the encoder output, $Cat$ denotes concatenation, $N$ refers to layer normalization, $C$ refers to 2D convolution, and $U$ refers to up-sampling. The concatenated input is passed through a *depth-wise (deep) convolution (DWC)* operation to extract the relevant salient features and reduce the channel dimension by a factor of two. The *DWC* operation is designed according to Equations 2 and 3 as follows:

$$DWC = D(R(C_2(R(C_1)))) \quad (2)$$

$$X' = DWC(X) \quad (3)$$

Here, $D$, $R$, and $C_i$ denote the dropout, ReLU activation function, and the number of 2D convolutions where $i = \{1, 2\}$, respectively. The result of DWC operation applied to $X$ is represented as $X'$. Then, we apply *multi-head self-attention (MSA)* [31] to the input. Normally, a position-based linear projection is used in *MSA* implementations, which requires more computations and also results in substantial loss of spatial details within images. Since precise image segmentation relies on these details, and the additional computational cost brought by linear projection must be alleviated, we use convolutional projection to compute *MSA* in this study. The use of convolutional projection reduces the number of parameters significantly, and provides more spatial context compared to linear projection. The number of attention heads for calculating *MSA* in the bottleneck layer and the other four decoder blocks is 10, 8, 6, 4, and 2, respectively. The implementation of *MSA* is shown through Equation 4. We then add the output of *MSA* with the input of *MSA* through skip-connections (Equation 4).

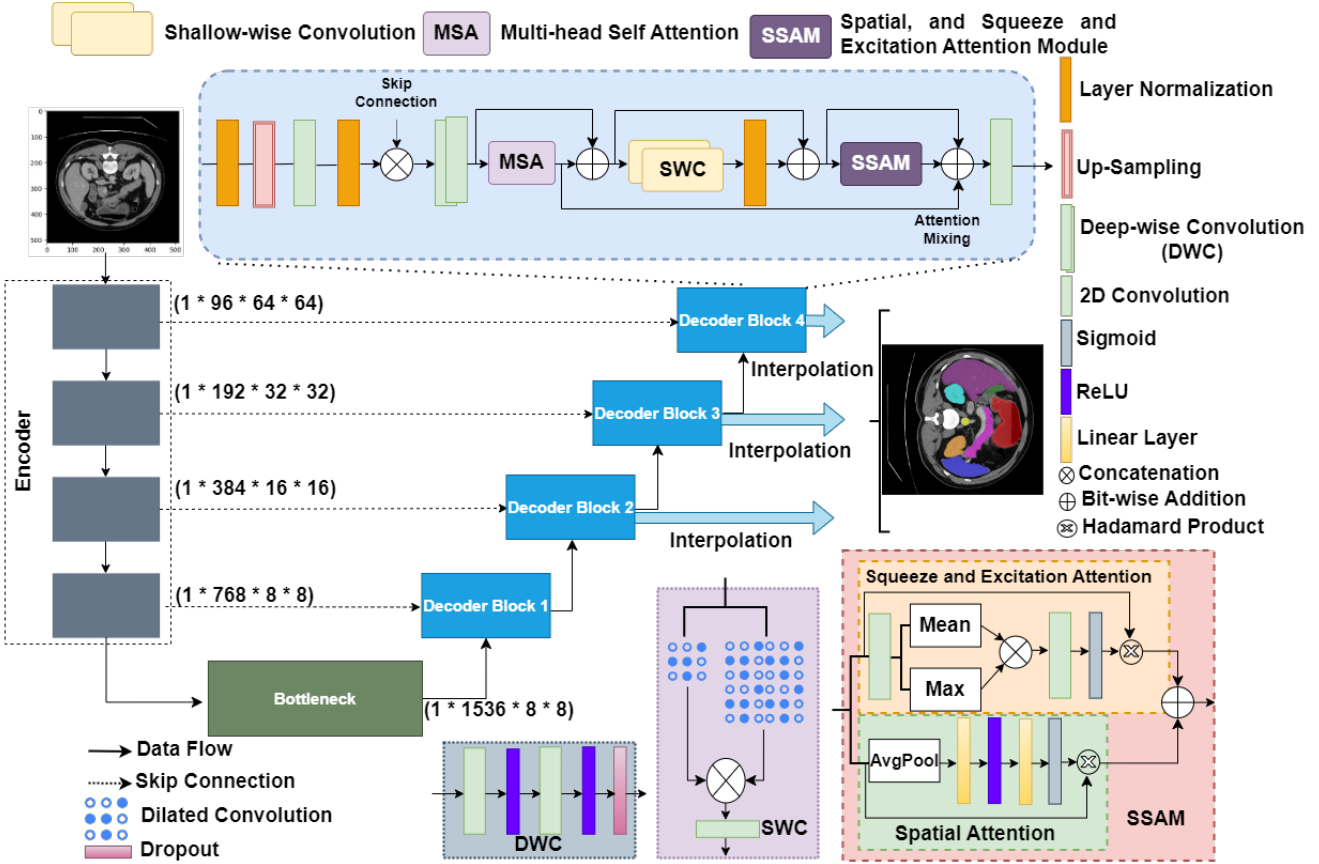

Figure 1. Medical Image Segmentation Transformer (MIST) with a novel Convolutional Attention Mixing (CAM) decoder for medical image segmentation. The left side of the network shows the encoder where we used pre-trained MaxViT, and the right side shows the decoder that generates the segmentation maps. Each block of the decoder was built by utilizing an attention-mixing strategy where we aggregated attentions computed at different stages. We used convolutional projected multi-head self-attention (MSA) instead of linear MSA to reduce computational cost and capture more salient features; and incorporated depth-wise (deep and shallow) convolutions (DWC and SWC) to extract more relevant semantic features and to increase kernel receptive field for better long-range dependency.

$$X'' = MSA(q,v,k) + X'; q,v,k = N(R(C_j(X'))) \quad (4)$$

where $N$ and $C_j$ represent the *layer-normalization* and *convolutional layer* for generating query ($q$), value ($v$) and key ($k$), i.e. $j = \{q, v, k\}$, respectively. The added features are passed through a *depth-wise (shallow) convolution (SWC)* to increase the kernel's receptive field and gain a better local and global spatial context of the image. Here, we aggregate and concatenate the features for better spatial gain and add them with the input through a skip connection, as presented in Equations 5-7.

$$SWC = C_1(Cat(X_1'', X_2'')) \quad (5)$$
$$X_1'' = R(C_2(X'', d=2)), X_2'' = R(C_3(X'', d=3)) \quad (6)$$
$$X''' = N(SWC(X'')) + X'' \quad (7)$$

where *Cat* denotes the concatenations, and $d$ refers to the dilation rate during convolution (to widen the kernel). Then, we apply the *spatial attention (SA)* and the *squeeze and excitation attention (SEA) module (SSAM)* over the features. The implementation of *SSAM* is shown in Equations 8-10 as follows:

$$SSAM = SA(X) * X + SEA(X) * X \quad (8)$$
$$SEA = \sigma_1(C_3(Cat(Mean(C_2(X)), Max(C_1(X))))) \quad (9)$$
$$SA = \sigma_2(L(R(L(P_a(X))))) \quad (10)$$

where $\sigma_i$, $L$, and $P_a$ denote Sigmoid activation function, linear layer, and adaptive average pooling, respectively. At the end, we perform attention mixing and aggregation. Here, *MSA attention*, *SSAM attention*, and *input to the SSAM module* are added together, as presented in Equation

11, allowing our CAM decoder to leverage the long-range dependencies among the image pixels in all spatial directions.
$$X'''' = SSAM(X''') + X''' + X'' \tag{11}$$
This implementation uses a 3*3 kernel size for all convolution operations. This study uses RGB image of size $3*256*256$ where height ($H$)= 256 and width ($W$)= 256. The decoder generates outputs of size $\{(96*2^i)*\frac{H}{2^{2+i}}*\frac{W}{2^{2+i}}\}$ in four different decoder blocks where $i = \{0, 1, 2, 3\}$, which are passed through a convolution to reduce the channel size to the number of classes by keeping the image size intact. We only use the last three outputs of the decoder for deep supervision as the region of interest will not be maintained in the low-resolution image. Then, we apply bilinear interpolation to resize the images to $H * W * N_{cls}$, where $N_{cls}$ is the number of classes, used as the final segmented maps of the decoder. The operation of the bottleneck layer is the same as the other four decoder blocks, except that the bottleneck gathers more information in the channel of the image. The number of channels has been increased twice, while the image size remains the same. Increasing channels through the bottleneck strengthens the encoder to retrieve more relevant salient features, which helps the decoder in better predicting the features map.

### 3.2. Multi-Axis Vision Transformer (MaxViT) Encoder

Recent advancements in medical image segmentation have shown potential for hierarchical transformers. One such transformer is MaxViT, which has demonstrated greater effectiveness in this field. In our MIST model, we have utilized the pre-trained MaxViT transformer as an encoder. Specifically, we have employed the MaxViT transformer with an image size of $3 * 256 * 256$, starting with two Stem blocks of width 32 and 64, followed by four stages with embedding sizes of 96, 192, 384, and 768, respectively. Each stage consists of a different number of MaxViT blocks (2, 2, 5, and 2, respectively), each composed of three separate blocks: Mobile Convolution Block (MBConv), which computes features and utilizes squeeze-and-excitation (SE) attention; Block Attention, which calculates block self-attention followed by a feed-forward network (FFN); and Grid Attention, which estimates grid self-attention followed by FFN, with all three blocks connected sequentially. Several skip connections are utilized inside each block to address the issue of vanishing gradients and facilitate information flow. MaxViT generates four feature representations of size $\{1 * (96 * 2^i) * \frac{H}{2^{2+i}} * \frac{W}{2^{2+i}}\}$ in four different stages where $i = \{0, 1, 2, 3\}$, which are passed to the CAM decoder through skip connections in this study.

### 3.3. Medical Image Segmentation Transformer (MIST)

The MIST model we have developed utilizes the MaxViT transformer as its encoder and CAM as its decoder. We feed RGB images through the encoder, and four feature maps (X1, X2, X3, and X4) in four different stages are generated, and concatenated with the up-sampled features of the decoder through the skip-connections. The extracted features of the last stage (X4) are also passed through a bottleneck layer, where the channel size has been increased twice to gather more information while the image size has been kept the same. The features from the bottleneck layer are propagated upward and up-sampled by a scale of two and concatenated with the features from the encoder. We consider the output $P_1$, $P_2$, and $P_3$ of decoder blocks 2, 3, and 4, respectively, for deep supervision, which we add together through the following equation (Equation 12) to generate the final segmentation maps and pass through Softmax to classify each pixel. We do not use the output of decoder block-1 as its image resolution is low, and the region of interest (ROI) is not maintained effectively.
$$y = P_1 + P_2 + P_3 \tag{12}$$

### 3.4. Loss Function

In this study, we have tested most of the proposed loss functions for medical image segmentation and found that the loss aggregation strategy by mixing features from multi-stage [27] is more effective in making the model better converge and efficient for image segmentation. According to the proposed loss function, we use $n$ prediction maps generated from our network. We generate $2^n - 1$ non-empty subsets of $n$ prediction maps and aggregate each non-empty subset to generate the final prediction map for loss calculation. This feature mixing strategy results in $2^n - 1$ individual prediction maps, including the original prediction maps generated by the model. Then, we apply *DICE* ($L_{DICE}$) and *CrossEntropy* ($L_{CE}$) loss functions to all $2^n - 1$ individual prediction maps and corresponding segmentation masks to estimate the losses. We aggregate all estimated losses using the following function to compute the final loss. Equation 13) serves as the ultimate loss function for our implementation.
$$L_{total} = \gamma L_{DICE} + (1 - \gamma) L_{CE} \tag{13}$$
where $\gamma = 0.3$ and $1 - \gamma = 0.7$ are the weights for $L_{DICE}$ and $L_{CE}$, respectively.

### 3.5. Datasets

This study uses the Automatic Cardiac Diagnosis Challenge (ACDC) and Synapse multi-organ datasets to evaluate the performance of our MIST architecture. The details of these datasets are as follows. The **ACDC dataset**[2] includes 100 cardiac MRI scans collected from different subjects, each with corresponding segmentation masks. Each MRI scan has multiple 2D slices featuring three organs - the right ventricle (RV), myocardium (Myo), and left ventricle (LV) - for segmentation. The dataset was divided into three sets: a training set consisting of 70 cases resulting in 1304 slices, a validation set containing 10 cases resulting in 182 slices, and a testing set with 20 cases providing 370 slices. The **Synapse multi-organ dataset**[3] has 30 abdominal CT scans and their corresponding segmentation masks, where each CT scan has around 85-198 2D slices adding up to 3779 axial contrast-enhanced abdominal slices. Each 2D slice of resolution 512*512 and voxel spatial resolution of [0:54-0:54] × [0:98-0:98]×[2:55:0])$mm^3$ has eight organs such as the aorta, gallbladder (GB), left kidney (KL), right kidney (KR), liver, pancreas (PC), spleen (SP), and stomach (SM), for segmentation. We randomly divide the dataset into two sets: a training set consisting of 18 CT scans and a testing set with the remaining 12 CT scans. We follow the same image preprocessing techniques outlined in the TransUNet transformer study for both datasets. To ensure a fair performance comparison like the TransUNet transformer study, we used the same CT and MRI scans for the training and testing sets.

---
[2]https://www.creatis.insa-lyon.fr/Challenge/acdc/

### 3.6. Evaluation Metrics

We evaluate our model's performance on the ACDC and Synapse multi-organ datasets using the DICE score and the 95% Hausdorff Distance (HD95) for the Synapse dataset. The DICE score and HD [27] are calculated according to the formulas of Equations 14 and 15:

$$DICE(X,Y) = \frac{2 * |X \cap Y|}{|X| + |Y|} \quad (14)$$

$$HD(X,Y) = \max\{d_{XY}, d_{YX}\}$$
$$= \max\left\{\max_{x \in X} \min_{y \in Y} d(x,y), \max_{y \in Y} \min_{x \in X} d(x,y)\right\} \quad (15)$$

Here, the X and Y denote the ground truth and segmented maps, respectively, and $d_{XY}$ and $d_{YX}$ represent the distances. HD95 is the $95^{th}$ percentile of the distances of the boundaries of X and Y.

### 3.7. Implementation

We have run the experiments on a Cluster Server, which is equipped with an NVIDIA A30 Tensor Core GPU with 4 nodes, and includes 64 cores, 512GB memory, and one A30 GPU (24GB) per node. We use PyTorch 1.13.1 and CUDA tools 11.2 for implementing the model and experiments. Pre-trained weights of MaxViT on the ImageNet dataset were used to initialize the MaxViT transformer, as an encoder. We train the model over both datasets for 300 epochs and record the best validation weights during training, to test our model's performance. This helps us to tackle the overfitting problem of our model during training. We use AdamW optimizer with a learning rate 0.0001 and weight decay of 0.0001. A weighted loss function is used by combining the CrossEntropy and DICE loss functions, wherein we mix the feature maps and aggregate the losses

---

[3] https://www.synapse.org/#!Synapse:syn3193805/wiki/217789

---

to compute the final loss. For both the ACDC and Synapse datasets, we resize images to 256*256 and keep the batch size of 12 for training. Additionally, we perform data augmentation techniques such as rotation, zooming, horizontal and vertical shifting, and flipping over both datasets. Some implementation components utilized in the MERIT study [27] have been incorporated here, as certain elements of the encoder model remain unchanged.

## 4. Results

Our MIST transformer with the novel CAM decoder was compared to state-of-the-art (SOTA) CNN and transformer models for medical image segmentation on the ACDC and Synapse multi-organ datasets. Our MIST transformer was found to be a new state-of-the-art model on both datasets. Details of the comparative results can be found in the following subsections. Figure 2 shows the ground truth and corresponding segmented images by our model.

### 4.1. Performance on ACDC dataset

MIST architecture's performance was compared to other recent SOTA models, such as SwinUNet, TransUNet, TransCASCADE, and Parallel MERIT. Our model outperformed all of them significantly. We conducted five experiments using the ACDC dataset and found that our model's average DICE score of 92.56±0.01 was higher than the highest average DICE score achieved by Parallel MERIT, which was 92.32. Our model is also more efficient in locating all small organs. The mean DICE scores for segmenting the right ventricle (RV), myocardium (Myo), and left ventricle (LV) were 91.23, 90.31, and 96.14, respectively, which were higher than the scores reported by the SOTA models. The comparative performance of our model over the ACDC dataset is shown in Table 1. These results indicate that our model, which uses the same encoder as Parallel MERIT but has a novel CAM decoder, is superior in accurately segmenting all three organs.

Table 1. Performance Comparison of Models on ACDC Dataset for organ segmentation. Organ abbreviations: RV (right ventricle), Myo (myocardium), LV (left ventricle). Mean DICE scores are over individual organ performance. The best results are highlighted in bold.

| Models | Mean DICE | RV | Myo | LV |
|---|---|---|---|---|
| TransUNet [6] | 89.71 | 88.86 | 84.53 | 95.73 |
| SwinUNet [3] | 90.00 | 88.55 | 85.62 | 95.83 |
| MT-UNet [32] | 90.43 | 86.64 | 89.04 | 95.62 |
| MISSFormer [12] | 90.86 | 89.55 | 88.04 | 94.99 |
| PVT-CASCADE [26] | 91.46 | 88.90 | 89.97 | 95.50 |
| nnUNet [13] | 91.61 | 90.24 | 89.24 | 95.36 |
| TransCASCADE [26] | 91.63 | 89.14 | 90.25 | 95.50 |
| nnFormer [42] | 91.78 | 90.22 | 89.53 | 95.59 |
| Parallel MERIT [27] | 92.32 | 90.87 | 90.00 | 96.08 |
| MIST (Proposed) | **92.56** | **91.23** | **90.31** | **96.14** |

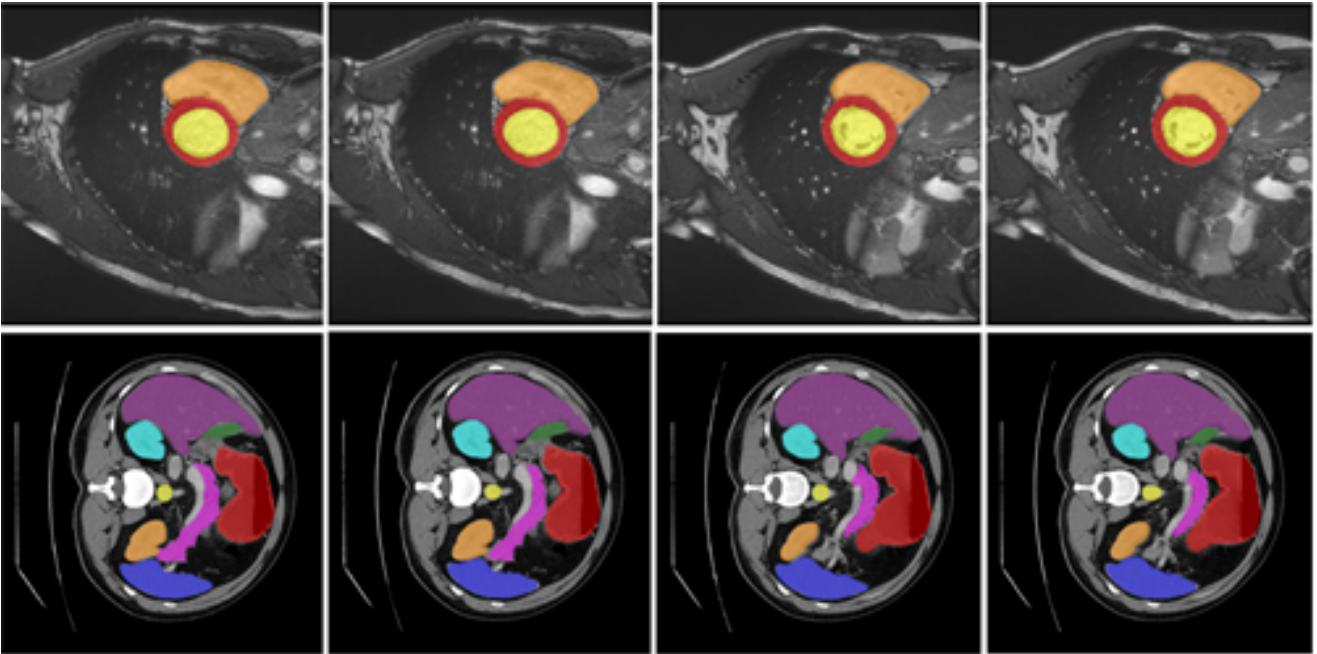

Figure 2. Segmentation qualitative outcomes. First row, ACDC dataset displays color-coded regions: Orange for right ventricle (RV), yellow for the left ventricle (LV), and red for myocardium. Second row represents Synapse dataset where the color codes are: yellow for the aorta, green for the gallbladder, orange for the left kidney, cyan for the right kidney, plum for the liver, pink for the pancreas, indigo for the spleen, and crimson for the stomach. The ground truth for different sample images is displayed in the first and third columns, while the corresponding segmented images are shown in the second and fourth columns.

Table 2. Comparative Analysis of Model Performance on Synapse Dataset for Organ Segmentation. Organ abbreviations: GB (gallbladder), KL (left kidney), KR (right kidney), PC (pancreas), SP (spleen), SM (stomach). Mean Dice scores represent individual organ performance. High DICE scores and low HD95 scores mean better performance. The best results are highlighted in bold.

| Models | Mean | | Aorta | GB | KL | KR | Liver | PC | SP | SM |
| --- | --- | --- | --- | --- | --- | --- | --- | --- | --- | --- |
| | DICE | HD95 | | | | | | | | |
| TransUNet [6] | 77.48 | 31.69 | 87.23 | 63.13 | 81.87 | 77.02 | 94.08 | 55.86 | 85.08 | 75.62 |
| SwinUNet [3] | 79.13 | 21.55 | 85.47 | 66.53 | 83.28 | 79.61 | 94.29 | 56.58 | 90.66 | 76.60 |
| MT-UNet [32] | 78.59 | 26.59 | 87.92 | 64.99 | 81.47 | 77.29 | 93.06 | 59.46 | 87.75 | 76.81 |
| MISSFormer [12] | 81.96 | 18.20 | 86.99 | 68.65 | 85.21 | 82.00 | 94.41 | 65.67 | 91.92 | 80.81 |
| PVT-CASCADE [26] | 81.06 | 20.23 | 83.01 | 70.59 | 82.23 | 80.37 | 94.08 | 64.43 | 90.1 | 83.69 |
| CASTformer [38] | 82.55 | 22.73 | 89.05 | 67.48 | 86.05 | 82.17 | **95.61** | 67.49 | 91.00 | 81.55 |
| TransCASCADE [26] | 82.68 | 17.34 | 86.63 | 68.48 | 87.66 | 84.56 | 94.43 | 65.33 | 90.79 | 83.52 |
| Parallel MERIT [27] | 84.22 | 16.51 | 88.38 | 73.48 | 87.21 | 84.31 | 95.06 | 69.97 | 91.21 | 84.15 |
| MIST (Proposed) | **86.92** | **11.07** | **89.15** | **74.58** | **93.28** | **92.54** | 94.94 | **72.43** | **92.83** | **87.23** |

### 4.2. Performance on Synapse Multi-Organ Dataset

We have conducted experiments on the Synapse dataset and the results are presented in Table 2. Here, the higher DICE score and lower HD95 score indicate better results. After conducting the experiments five times, our model achieved a mean DICE score of 86.95±0.08 and a mean HD95 of 11.08±0.07. These scores are over 4% and 2% better, respectively, compared to the recent TransCASCADE and Parallel MERIT models. Our model outperforms Parallel MERIT with a significant margin in segmenting all the organs. These comparative results demonstrate that our novel CAM decoder combined with the MaxViT encoder is performing well, indicated by the significant performance improvment compared to the SOTA Parallel MERIT model, which shares the same encoder but uses different decoders. The reason for our model's improved per-formance can be attributed to the mixing and aggregation of three different attention mechanisms computed at different stages of the decoder block. This allows our model to capture each pixel's local and global context in all spatial dimensions. In addition, we computed convolutional projected multi-head self-attention (MSA) to enable capturing long-range dependencies of each pixel, as well as more spatial pixel information compared to the linear projected MSA. We also used shallow convolutions (SWC), where different dilated convolutions were used to capture salient information and increase the kernel receptive field, which may have contributed to our model's better performance.

Table 3. Comparison of Ablation Studies: The ideal configuration is highlighted in bold. In the remaining experiments, only a single parameter was altered while keeping the others constant to observe the impact of each parameter on the performance.

| Attention-mixing | MSA Projection | Dilations | Attention-aggregation | Mean DICE | HD95 |
|---|---|---|---|---|---|
| Yes | Linear | 2, 3 | Concatenation | 84.87 | 13.15 |
| Yes | Convolution | 1, 2 | Concatenation | 86.23 | 12.65 |
| Yes | Convolution | 2, 3 | Summation | 85.43 | 12.01 |
| No | Convolution | 2, 3 | Concatenation | 83.81 | 14.11 |
| **Yes** | **Convolution** | **2, 3** | **Concatenation** | **86.95** | **11.08** |

### 4.3. Ablation Experiments

We analyzed the effects of attention-mixing, two projections for MSA, varying number of dilations in the shallow convolution (SWC) module, and attention-aggregation techniques in SWC on the segmentation performance. We conducted tests on the Synapse multi-organ dataset to evaluate the effects of various settings. As part of our analysis, we experimented with different parameters such as attention-mixing (yes or no), MSA projection (convolution or linear), dilations (1 and 2 or 2 and 3), and attention-aggregation (concatenation or summation). We tried all 16 possible combinations and discovered that the best accuracy was achieved with attention-mixing set to yes, MSA projection set to convolution, dilations set to 2, 3, and attention-aggregation set to concatenation. To demonstrate the impact of each parameter, the experimental results are shown in Table 3, obtained by keeping the other settings constant at their optimal values. The reported results show that the attention-mixing strategy significantly impacted capturing both local and global contexts, resulting in a significant increase in segmentation performance by around 3% of our MIST model compared to when our model was designed without attention-mixing. We have significantly improved the performance by using convolutional projected MSA instead of linear projected MSA. Our experiments have demonstrated that it has reduced computational costs by approximately 11% while increasing segmentation accuracy by around 2%. The convolutional projection replaces the linear projection, requiring significantly fewer parameters to be learned. The number of parameters with linear projection was 107.23M which reduced to 95.74M (11% lower) when using convolutional projection. Additionally, convolution's inherent local nature helps capture more spatial information. We use depth-wise (deep and shallow) convolutions to increase the receptive field of kernels in our CAM decoder. Through experimentation, we have found that using 2 and 3 dilations for two different convolutions works best. We have also compared the effects of concatenation versus summation of feature maps from two dilated convolutions on our model's performance where concatenation of feature maps yields around 1.5% improved performance compared to the summation of feature maps. However, if we use concatenation, we can retain all the information, even though it may result in higher computational costs. Based on our comparisons, it is evident that implementing the attention-mixing strategy, convolutional projected MSA, SWC with dilations of 2 and 3, and concatenation of feature maps from two different dilated convolutions, can significantly enhance accuracy and establish the MIST as a new SOTA model for medical image segmentation.

### 5. Conclusion

We proposed a Medical Image Segmentation Transformer (MIST) incorporating a novel Convolutional Attention Mixing (CAM) decoder for medical image segmentation. We introduced an attention-mixing strategy in our newly proposed decoder where we aggregated convolutional multi-head self-attention (MSA), spatial, and squeeze and excitation attentions, which were computed at different stages of each decoder block, enabling our model to capture all forms of dependencies among pixels effectively. We found that incorporating convolutional MSA instead of linear MSA yielded improved accuracy and significantly reduced computational costs. We utilized depth-wise (deep and shallow) convolutions to capture more salient information and increase the receptive field of kernels, respectively. Using shallow convolutions with different dilations helped our model to leverage the pixel's long-range dependencies more efficiently. In addition, by incorporating multi-stage features from the encoder to the decoder through skip connections, we improved information flow and preserved relevant semantic information for better segmentation. Our model outperformed other models on both

ACDC and Synapse multi-organ datasets, achieving new SOTA in mean DICE and mean HD95 scores. Our MIST model with CAM decoder has the potential to be effective in segmenting other medical image datasets as well.